\pdfoutput=1

\documentclass[11pt]{article}

\usepackage[]{EACL2023}

\usepackage{times}
\usepackage{latexsym}
\usepackage{todonotes}
\usepackage[T1]{fontenc}

\usepackage[utf8]{inputenc}

\usepackage{microtype}

\usepackage{inconsolata}
\usepackage{booktabs}
\usepackage{comment}
\usepackage{amsmath}
\usepackage{multirow}
\usepackage{xcolor}
\usepackage{colortbl}
\usepackage{etoolbox}
\usepackage{pgf}
\usepackage{xspace}

\newcommand{\ie}{\textit{i.e., }\xspace}

\usepackage{url}

\newcommand{\GenScore}{\textit{$\text{proba}_G$}\xspace}
\newcommand{\OccScore}{\textit{$\text{score}_O$}\xspace}

\usepackage{footnote}

\definecolor{high}{HTML}{E8E3E3}
\definecolor{low}{HTML}{b43216}
\def\cca#1{
    \ifdimcomp{#1pt}{>}{48.88 pt}{#1}{%
    \ifdimcomp{#1pt}{<}{0.00 pt}{#1}{%
         \pgfmathparse{int(round(100*(#1/(48.88-0.0))-(0.0*(100/(48.88-0.0)))))}%
        \xdef\tempa{\pgfmathresult}%
        \cellcolor{high!\tempa!low!90} #1%
    }}
}

\title{Measuring Normative and Descriptive Biases in Language Models \\ Using Census Data}

\author{Samia Touileb \\
  University of Bergen \\
  \texttt{samia.touileb@uib.no} \\\And
  Lilja Øvrelid\\
  University of Oslo \\
  \texttt{liljao@ifi.uio.no} \\ \And
  Erik Velldal\\
  University of Oslo \\
  \texttt{erikve@ifi.uio.no} \\}

\begin{document}
\maketitle
\begin{abstract}

We investigate in this paper how distributions of occupations with respect to gender is reflected in pre-trained language models. 
Such distributions are not always aligned to normative ideals, nor do they necessarily reflect a descriptive assessment of reality. In this paper, we introduce an approach for measuring to what degree pre-trained language models are aligned to normative and descriptive occupational distributions. To this end, we use official demographic information about gender--occupation distributions provided by the national statistics agencies of France, Norway, United Kingdom, and the United States. We manually generate template-based sentences combining gendered pronouns and nouns with occupations, and subsequently probe a selection of ten language models covering the English, French, and Norwegian languages. The scoring system we introduce in this work is language independent, and can be used on any combination of template-based sentences, occupations, and languages. The approach could also be extended to other dimensions of national census data and other demographic variables. 
\end{abstract}

\section{Introduction}

Pre-trained language models (LMs) may contain various types of biases, and the field of NLP has seen a lot of work in recent years on attempting to identify, mitigate, and reduce these biases. Biases can originate both from the unlabeled texts used for pre-training these LMs, and from texts and annotations used for tuning downstream classifiers. LMs have become a cornerstone in most NLP model architectures, and the extent to which they reflect, amplify, and spread the biases present in their training data is still a problematic issue to be solved. 

Several efforts in this direction have focused on \textit{gender} as a variable \cite{touileb-nozza-2022-measuring,touileb-etal-2021-using,ousidhoum-etal-2021-probing,nozza-etal-2021-honest,touileb-etal-2020-gender,saunders-byrne-2020-addressing,bhaskaran-bhallamudi-2019-good, cho-etal-2019-measuring, pratesetal2018}, also in correlation with occupations \cite{borchers-etal-2022-looking,touileb-etal-2022-occupational,bolukbasi2016man}. While there have been several efforts on exploring the existing biases related to these demographic variables, most work approaches the task from a normative point of view \cite{blodgett2021sociolinguistically}, where equality between the demographic distributions is prioritized.

Although normativity in this aspect is crucial for certain applications, we argue that it is also interesting to explore the task from a descriptive perspective. This is especially interesting for occupations, since a descriptive and realistic view of society already contains gender disparities. We propose that national census data, in our case about gender-occupation distributions, can offer a reliable ground truth against which model predictions can be compared. Moreover, we argue for taking both normative and descriptive assessments into account, in order to give a broader picture of the representations of demographics within LMs. This has also been partly pointed out by \citet{blodgett2020language}, who stress the importance of the connection between language and social hierarchies, which has not been taken into consideration in most previous work on bias in NLP.  

In this paper, we introduce a new score for measuring how LMs are aligned with normative and descriptive occupational demographic distributions. We use demographic distributions covering occupations in four countries, namely France, Norway, the United Kingdom, and the United States. We manually select gendered pronouns and nouns, as well as specific verb phrases, to construct template-based sentences, subsequently used to probe a selection of ten LMs covering the relevant languages.

Our contributions include; (i) creating novel benchmark datasets for English, French, and Norwegian based on manually crafted templates to measure occupational gender biases, (ii) proposing a scoring system to measure normative and descriptive biases in LMs, and (iii) releasing our code and data for reproducibility. 

In what follows, we give a detailed description of our new benchmark datasets in Section~\ref{sec:data}. We then, in Section~\ref{sec:scores}, give a detailed description of the normative and descriptive bias scores, and present our analysis on ten LMs as proof of concept. We discuss and summarize our findings in Section~\ref{sec:results}, and conclude by discussing possible directions for future work in Section~\ref{sec:conclusion}. The limitations of our work are discussed in the Limitations Section.

\section{Benchmark datasets}\label{sec:data}

In this work we develop a set of benchmark templates for English, French, and Norwegian that cover occupations in France, Norway, the United Kingdom (UK), and the United States (US). These templates are then used for probing different LMs. More details are given in what follows.

\paragraph{Occupations}

We retrieve country-specific lists of occupations and their associated (male/female) gender ratios from the national statistics bureaus of France, Norway, UK, and the US.\footnote{All of the statistics were retrieved in October 2022.} 
This resulted in 235 occupations from France,\footnote{\url{https://dares.travail-emploi.gouv.fr/donnees/portraits-statistiques-des-metiers}} 415 from Norway,\footnote{\url{https://utdanning.no/likestilling}} 325 from the UK,\footnote{\url{https://www.nomisweb.co.uk/datasets/aps168/}} and finally 314 occupations from the US.\footnote{\url{https://www.bls.gov/cps/cpsaat11.htm}} All of these occupations were listed in either masculine singular or masculine plural form. As some of the languages we are focusing on inflect nouns for gender, we manually generate for each occupation in singular masculine form the corresponding forms in singular feminine, plural feminine, and plural masculine. This was performed by a native speaker of Norwegian and French, and a proficient speaker of English. Table \ref{tab:example_occupations} shows the top 5 female-dominated, male-dominated, and gender balanced occupations in each census data.

\begin{table*}[t]
    \centering
    \small
     \begin{tabular}{l|l|l|l}
    \toprule

    & Female-dominated occupations & Male-dominated occupations & Gender-balanced occupations\\
    \midrule

    \multirow{5}{*}{FR} 
    & midwife & navy officer and boatswain & doctor \\ 
    & kindergarten assistant & construction machinery operator & higher education teacher \\
    & secretary & pipe fitter & medical device specialist \\
    & office secretary & panel beater & admin. 
    and financial executive \\ 
    & executive secretary & carpenter & dentist \\

    \midrule

    \multirow{3}{*}{NO}
    & knitter & coastal skipper & doctor \\ 
    & midwife & chief engineer & architect \\ 
    & public health nurse & scaffold builder  & lawyer\\ 
    & skin care specialist &  roofer & politician\\ 
    & dental health secretary &  bricklayer & associate professor \\

    \midrule

    \multirow{5}{*}{UK} 
    & midwife & roofer, roof tiler and slater & barrister and judge \\ 
    & school secretary & carpenter and joiner & laboratory technician \\ 
    & dancer and choreographer & construction and building 
    supervisor & paramedic\\ 
    & dental nurse & bricklayer and mason & 
    industrial trainer and instructor \\ 
    & medical secretary & vehicle technician and  mechanic 
    & legal professional  \\

    \midrule

    \multirow{5}{*}{US}
    & skincare specialist & cement mason 
    & insurance sales agent \\
    & preschool and kindergarten teacher & electrical power-line installer 
    & medical scientist \\
    & executive secretary 
    & crane and tower operator & dental 
    laboratory technician 
    \\
    & speech-language pathologist & heavy vehicle 
    technician and mechanic & photographer \\
    & dental hygienist & bus and truck mechanic 
    & advertising sales agent \\
    \bottomrule
    \end{tabular}
    \caption{Top 5 gender-dominated and gender-balanced occupations in census data from France (FR), Norway (NO), the United Kingdom (UK), and the United States (US). The occupations presented here are either dominated by more than 98\% of either gender, or have a more balanced distribution (between 45\% and 55\%) between both female and male genders.}
    \label{tab:example_occupations}
\end{table*}

\paragraph{Templates} 

Our work builds on the methodology of template-based probing. To measure a model's occupational biases we follow the same procedure for all languages. Our templates are based on the gender-inflected occupations, preceded by a sequence of selected gendered pronouns and a set of gender-specific identifier terms in singular and plural forms, followed by  
a predicate generically denoting the act of having an occupation.    
As an  example, a template could be: 
\[\underbrace{\text{The woman}}_{\text{gender-specific identifier}} \underbrace{\text{ worked as a }}_{\text{predicate}} \underbrace{\text{ nurse}}_{\text{occupation}}\]

We select 28 gender-specific identifiers, and 6 predicates for all three languages. The full list of gender-specific identifiers can be found in Table~\ref{tab:identifiers_templates} and the list of predicates in Table  \ref{tab:predicates_templates} in Appendix~\ref{sec:appendix}. Combining these identifiers and predicates with our country-specific occupations, we create a set of 12.726 template-based probes for French occupations, 69.720 for Norwegian, 50.700 for the UK, and 48.984 for the US.

The templates we created cover different grammatical tenses, such that each template is given in the past, present, and future tense. 
We have decided to include such a broad collection of variations to the templates to get a better representation of how occupations are correlated with genders, especially since research has shown that bias probes are sensitive to grammatical tense \cite{touileb-2022-exploring}.

\begin{table}[t]
    \centering
    \small
    \begin{tabular}{@{}lll@{}}
    \toprule
    Norwegian & English & French \\
\midrule 
Brødrene & He & Elle \\
Broren & She & Elles \\
Dama/Damen & The aunt & Il \\
Damene & The aunts & Ils \\
Datteren & The boy & L'homme \\
Døtrene & The boys & L'oncle \\
Faren & The brother & La dame \\
Fedrene & The brothers & La femme \\
Gutten & The daughter & La fille \\
Guttene & The daughters & La mère \\
Han & The father & La soeur \\
Hun & The fathers & La tante \\
Jenta/Jenten & The girl & Le fils \\
Jentene & The girls & Le frère \\
Kvinnen & The ladies & Le garçon \\
Kvinnene & The lady & Le père \\
Mannen & The man & Les dames \\
Mennene & The men & Les femmes \\
Mødrene & The mother & Les filles \\
Moren & The mothers & Les fils \\
Onkelen & The sister & Les frères \\
Onklene & The sisters & Les garçons \\
Sønnen & The son & Les hommes \\
Sønnene & The sons & Les mères \\
Søsteren & The uncle & Les oncles \\
Søstrene & The uncles & Les pères \\
Tanten & The woman & Les sœurs \\
Tantene & The women & Les tantes \\
    \bottomrule
    \end{tabular}
    \caption{Gender-specific pronouns and identifiers.}
    \label{tab:identifiers_templates}
\end{table}

\begin{table}[t]
    \centering
    \small
    \begin{tabular}{@{}lrr@{}}
    \toprule
     Model & Normative & Descriptive  \\
\midrule 
    NorBERT            & \cca{16.23} & \cca{39.31} \\
    NorBERT2           & \cca{3.17}  & \cca{34.67} \\
    NB-BERT            & \cca{18.55} & \cca{36.50} \\
    NB-BERT\_Large     & \cca{11.35} & \cca{40.90} \\
    BERT\_UK           & \cca{18.05} & \cca{35.33} \\
    BERT\_large\_UK    & \cca{13.73} & \cca{40.43} \\
    RoBERTa\_base\_UK  & \cca{0.15}  & \cca{34.56} \\
    RoBERTa\_large\_UK & \cca{0.00}  & \cca{34.56} \\
    BERT\_US           & \cca{17.25} & \cca{43.29} \\
    BERT\_Large\_US    & \cca{12.46} & \cca{48.88} \\
    RoBERTa\_base\_US  & \cca{0.15}  & \cca{42.81} \\ 
    RoBERTa\_Large\_US & \cca{0.31}  & \cca{42.81} \\
    CamemBERT          & \cca{10.46} & \cca{34.10} \\ 
    BARThez            & \cca{6.45}  & \cca{37.08} \\
    \bottomrule
    \end{tabular}
    \caption{Normative and descriptive occupational bias scores.}
    \label{tab:normative_desc_scores}
\end{table}

\section{Method}\label{sec:scores}

LMs trained with a masked language modelling objective are trained such that random tokens in the input training data are replaced with a placeholder token, \texttt{[MASK]}, which will subsequently be predicted by the trained model. Template-based approaches to probe biases take advantage of this feature of LMs. For our purposes, we mask the gendered identifier in each template-generated probe (as introduced in Section \ref{sec:data}), and use the returned probability of each masked identifier to compute our bias scores. A masked version of the example template above would be: \[\underbrace{\text{The [MASK]}}_{\text{gender-specific identifier}} \underbrace{\text{ worked as a }}_{\text{predicate}} \underbrace{\text{ nurse}}_{\text{occupation}}\]

\paragraph{Language models}

We select ten LMs covering the three languages English, French, and Norwegian. All models are available from the HuggingFace library \cite{wolf2020transformers}.
We use four Norwegian models, four English models, and two French models. These are: 
\begin{itemize}\itemsep2pt
    \item NorBERT~\cite{kutuzov2021large}: trained from scratch on the Norwegian newspaper corpus\footnote{\url{https://www.nb.no/sprakbanken/ressurskatalog/oai-nb-no-sbr-4/}}, and Norwegian Wikipedia. The model comprises about two billion word tokens. 
    \item NorBERT2\footnote{\url{https://huggingface.co/ltgoslo/norbert2}}: the non-copyrighted subset of the Norwegian Colossal Corpus (NCC)\footnote{\url{https://github.com/NbAiLab/notram/blob/master/guides/corpus_description.md}} and the Norwegian subset of the C4 web-crawled corpus \cite{xue-etal-2021-mt5} were used to train this model from scratch. It comprises about 15 billion word tokens.  
    \item NB-BERT\_base~\cite{Kummervold2021}: trained on the full version of the NCC corpus. This model used the architecture of the BERT cased multilingual model \cite{devlin2019}. It  comprises around 18.5 billion word tokens. 
    \item NB-BERT\_Large\footnote{\url{https://huggingface.co/NbAiLab/nb-bert-large}}: trained similarly to the NB-BERT\_base model.
    \item BERT\_base~\cite{devlin2019} and BERT\_Large: trained on English Wikipedia and Google’s Books Corpus.
    \item RoBERTa\_base~\cite{liuetal2019} and RoBERTa\_Large: trained on the BookCorpus, English Wikipedia, CC-news corpus (English news), OpenWebText dataset, and Stories dataset (a subset of the Common Crwal corpus).
    \item CamemBERT~\cite{martin2020camembert}: trained on the OSCAR corpus \cite{OrtizSuarezSagotRomary2019}, which is a multilingual corpus created by filtering the Common Crawl corpus.
    \item Barthez~\cite{eddine2020barthez}: trained on the French part of the Common Crawl and Wikipedia, in addition to various smaller corpora \cite{eddine2020barthez}.
\end{itemize}

\paragraph{Scoring system}

The scoring system we introduce is the same for both bias scores. We will give more details on the differences of the scores in their respective sections. 

For each template, and for each language, 28 gender-specific identifiers and 6 different predicates were used with each occupation. To compute the scores, we average over the gendered-identifiers and the predicates of the LMs' returned probabilities for each template. For each template, only one gender is represented (female or male). If the language inflects for gender, all components of a template reflect the gender in question, otherwise it is  only reflected in the identifier. 

We average the scores for a given occupation by gender, by summing and normalizing the probabilities of each identifier and the total probability values returned by the LM, here dubbed \GenScore, where \textit{G} can be female or male (equation (1)). Then using this overall probability of a gender for a template, we average these values over all templates related to the occupation (equation (2)). More formally, for a language model \textit{LM}, for each occupation \textit{O}, there are a number of templates \textit{T}, and a number of identifiers \textit{i} and predicates \textit{p}, reflecting a gender \textit{G}. We define the bias score as follows: 

\begin{equation}
\GenScore = \frac{\sum_{i}T_{p}}{|i|}
\end{equation}

\begin{equation}
\OccScore = \frac{\sum_{proba_G} T_O}{|T_O|}    
\end{equation}

\paragraph{Descriptive bias score}

Once the scores \OccScore are computed, the descriptive bias score compares the percentages of distribution of occupations in the LMs to the ground truth data that comes from the respective census data of our countries of interest. We impose a threshold on the gender distributions in such a way that the category of \textit{gender-imbalanced} occupations here corresponds to all occupations exceeding 55\% of distribution for one gender, while \textit{gender-balanced} occupations are those which percentages lie around 50\%$_{\pm5}$ for each gender. 

We look at the extent to which this score aligns with the census data. We compute an overall score disregarding gender, in addition to class-level scores: female dominated occupations (more than 55\% in census are females), male dominated occupations (more than 55\% in census are males), neutral occupations (between 45\% and 55\% of occupations in census for either gender).

\paragraph{Normative bias score}

The normative bias score also builds on top of the scores \OccScore, and compares the resulting distribution of occupations in LMs to a normative description of all occupations, such that percentages of either gender should be around 50\%$_{\pm5}$. 

From a normative point of view, equal representations should be given to females and males. Instead of just setting the distribution to a strict value of 50-50, we decided that for either gender, the distribution should range anywhere between 45\% and 55\% in the census data. This to say, that if an occupation has 45\% and 55\% males, we consider it a balanced distribution. 

\section{Results and discussion}\label{sec:results}

\begin{table}[t]
    \centering
    \small
    \begin{tabular}{@{}lrrrr@{}}
    \toprule
 Model & Neutral & Female & Male \\
\midrule 
NorBERT            & \cca{1.46} & \cca{22.34} & \cca{15.50}\\
NorBERT2           & \cca{0.24} & \cca{33.57} &  \cca{0.85}\\
NB-BERT            & \cca{1.46} & \cca{23.68} & \cca{11.35}\\
NB-BERT\_Large     & \cca{0.12} & \cca{33.82} &  \cca{6.95}\\
BERT\_UK           & \cca{1.54} & \cca{33.02} &  \cca{0.77}\\
BERT\_Large\_UK    & \cca{1.23} & \cca{31.63} &  \cca{7.56}\\ 
RoBERTa\_base\_UK  & \cca{0.00} & \cca{34.56} &  \cca{0.00}\\
RoBERTa\_Large\_UK & \cca{0.00} & \cca{34.56} &  \cca{0.00}\\
BERT\_US           & \cca{2.39} & \cca{39.93} &  \cca{0.95}\\
BERT\_Large\_US.   & \cca{1.75} & \cca{40.09} &  \cca{7.02}\\ 
RoBERTa\_base\_US   & \cca{0.00} & \cca{42.81} &  \cca{0.00}\\
RoBERTa\_Large\_US & \cca{0.00} & \cca{42.81} &  \cca{0.00}\\
CamemBERT          & \cca{0.00} &  \cca{0.00} & \cca{34.10}\\ 
BARThez            & \cca{0.00} &  \cca{0.00} & \cca{37.08}\\
\bottomrule
    \end{tabular}
    \caption{Descriptive bias scores of gender-imbalanced and gender-neutral occupations. The two gender-imbalanced occupations cover female dominated occupations (more than 55\% in census are females), and male dominated occupations (more than 55\% in census are males). The gender-neutral occupations are those with distributions between 45\% and 55\% in census data for either gender.
    }
    \label{tab:scores_fine}
\end{table}

Table~\ref{tab:normative_desc_scores} shows the resulting normative and descriptive bias scores of the ten LMs. All scores represent percentages, \ie the percentage of model predictions that align with our normative values or descriptive demographic distributions. With no surprise, it is clear that all models exhibit fairly weak performance according to the normative bias score. The weakest performing model normatively speaking is RoBERTa (both base and large) on both UK and US statistics. BERT seems to be a bit better on UK statistics, but the difference is not significant. For the remaining languages, NorBERT2 is the worst Norwegian model normatively and BARThez is the worst of the two French models. 

Results of the descriptive scores are in general higher. Most models seem to reflect the demographic occupational distribution to a certain extent. Both BERT models achieve highest descriptive scores, performing best on the US census data. 
While the RoBERTa models obtain the lowest performance in terms of the normative score, they rank second in the descriptive score on the US census data. NorBERT2 is still the weakest performing Norwegian model, ranking last both descriptively and normatively, while BARThez seems to yield the best descriptive score for French.

To get a more detailed overview of which types of occupations the tested models seem to represent the best, we also computed the descriptive scores of gender-imbalanced and gender-neutral occupations separately. Results can be seen in Table~\ref{tab:scores_fine}.
Interestingly, all Norwegian and English models are better at identifying female-dominated occupations, while the two French models seem to only identify male-dominated occupations.

All models exhibit the lowest scores on gender-neutral occupations, hinting at the tendency that models correlate most occupations with one gender, rather then equally representing them. This would also align with the lower normative scores that we generally see. 

Since the occupations in the census data differ from country to another, it is difficult to compare and rank models across languages. A fair comparison of these models is to focus on performance by country rather than across them. Even if we state that some models are the best or worst using one scoring system, the country-level scores are the most important measure of bias in the models.  

\section{Conclusion}\label{sec:conclusion}

We have introduced a new scoring system for measuring occupational biases in pre-trained language models. The scoring system allows the attribution of two scores: a normative score and a descriptive score. While the normative score sheds light on to what extent the correlations between genders and occupations are balanced, the descriptive score uses real world demographic distributions to reflect to what degree the language models reflect reality. 

As a proof of concept we test our scoring systems on ten language models covering the French, Norwegian, and English languages. It comes as no surprise that all models exhibit low scores when using the normative scoring, while most of them have an adequate score when measured descriptively. What is more interesting is that our scoring mechanism allows us to separate between the normative and descriptive aspects of the model properties. All templates and codes are made publicly available on our GitHub repository.\footnote{\url{https://github.com/SamiaTouileb/Normative-Descrptive-scores}} 

While we have limited our analysis to three languages, our approach is language agnostic and only requires language specific templates and demographic statistics on the distributions of occupations with respect to gender, something most national census agencies should be able to provide. Moreover, the approach could also be extended to other  dimensions of national census data and other demographic variables.  

On the note of future directions, we also plan to investigate cross-cultural effects, by comparing models for different languages across the gender--occupation lists from different nations, and also including multi-lingual models. Moreover, we also plan to systematically test the impact of different text sources (used to train language models) on our bias scores.  

\section*{Limitations}\label{sec:limitations}
The major limitation of our work is that we focus on a binary gender setting. We acknowledge the fact that gender as an identity spans more than just two categories, however, the demographic census data we use have only the two genders (female and male) represented.

As proof of concept, the templates we use in this work are limited to one framing of how gendered pronouns and nouns can co-occur with occupations. Extending this to more diverse templates might give a broader context and a better representation of genders in LMs.

The applicability of English models to the UK and US census data, and the French models to the French census data, might also give a skewed representation of occupations. Both English and French are spoken across many countries, which might have an effect on the representation of occupations and genders in the language models. It would therefore be interesting to investigate to what extent language-specific language models reflect census data from different countries. 



\section*{Acknowledgements}
This work was supported by industry partners and the Research Council of Norway (RCN) through funding to \textit{MediaFutures: Research Centre for Responsible Media Technology and Innovation}, through the RCN's Centers for Research-based Innovation scheme, project number 309339.

\bibliography{anthology,custom}
\bibliographystyle{acl_natbib}

\appendix

\section{Appendix}
\label{sec:appendix}

\begin{table}[h]
    \centering
    \begin{tabular}{@{}l|l@{}}
    \toprule
    Language & Predicates \\
    \midrule
    
    \multirow{6}{*}{Norwegian} 
    & jobber som \\ 
    & jobbet som \\ 
    & skal jobbe som \\ 
    & vil jobbe som \\ 
    & ville jobbe som \\ 
    & kommer til å jobbe som \\ 

    \midrule
    \multirow{12}{*}{English}
    & are going to work as \\ 
    & is going to work as a \\ 
    & want to work as \\ 
    & wanted to work as \\ 
    & wanted to work as a \\ 
    & wants to work as a \\ 
    & will work as \\ 
    & will work as a \\ 
    & work as \\ 
    & worked as \\ 
    & worked as a \\ 
    & works as a \\ 

    \midrule
    \multirow{12}{*}{French} 
    & est \\ 
    & étaient \\ 
    & était \\ 
    & sera  \\ 
    & serons \\ 
    & sont \\ 
    & va travailler comme \\ 
    & veulent être \\ 
    & veux être \\ 
    & vont travailler comme \\ 
    & voulaient être \\ 
    & voulais être \\ 
    \bottomrule
    \end{tabular}
    \caption{Language-specific predicates.}
    \label{tab:predicates_templates}
\end{table}

\end{document}